\documentclass{article}

\usepackage[preprint]{aaj2026}

\usepackage[utf8]{inputenc} 
\usepackage[T1]{fontenc}    
\usepackage{hyperref}       
\usepackage{url}            
\usepackage{booktabs}       
\usepackage{amsfonts}       
\usepackage{nicefrac}       
\usepackage{microtype}      
\usepackage{xcolor}         
\usepackage{graphicx}
\usepackage{subcaption}
\usepackage{amsmath}

\title{Vision-Based Water Level and Flow Estimation}

\author{%
    ZhiXin Sun\\
  PowerChina Zhongnan Engineering Corporation Limited \\
  \texttt{sunzxjdi@gmail.com} 
}

\begin{document}

\maketitle

\begin{abstract}
With the rapid evolution of computer vision, vision-based methodologies for water level and river surface velocity estimation have reached significant maturity. Compared to traditional sensing, these techniques offer superior interpretability, automated data archiving, and enhanced system robustness. However, challenges such as environmental sensitivity, limited precision, and complex site calibration persist. This work proposes an integrated framework that synergizes state-of-the-art (SOTA) vision models with statistical modeling. By leveraging physical priors and robust filtering strategies, we improve the accuracy of water level detection and flow estimation. Code will be available at \url{https://github.com/sunzx97/Vision_Based_Water_Level_and_Flow_Estimation.git}
\end{abstract}

\section{Introduction}
\textbf{Vision-based water level recognition:}
With the advancement of semantic segmentation techniques, pixel-level water level recognition has become feasible through semantic segmentation methods \cite{app13095614}. However, conventional approaches typically require collecting and annotating site-specific images to train a dedicated segmentation model for each monitoring station. To address this limitation, this paper proposes a prior-knowledge-based semantic segmentation framework for water level recognition. Unlike traditional semantic segmentation approaches, the proposed method eliminates the need for site-specific training data.

Instead, the method exploits a physical prior in natural river scenes: water is consistently distributed below the riverbank. After projection through the imaging process, the lower boundary of the water region generally remains distinguishable and geometrically stable. Based on this observation, interactive segmentation models, such as RITM\cite{reviving2021} and SAM2\cite{ravi2024sam2}, are employed by placing simulated prompt points within the water region to guide segmentation.

Furthermore, to suppress erroneous segmentation results, a segmentation quality filtering strategy is introduced. In addition to extracting the water body, the proposed framework incorporates physical constraints derived from shoreline geometry. Since the shoreline structure at a fixed monitoring station is relatively stable over time, it can be utilized as a prior for evaluating segmentation reliability. An iterative refinement process is performed, and segmentation results that fail to satisfy a predefined confidence threshold after multiple iterations are discarded.

Once the water region has been segmented, a relative water level can be estimated from the image. To further convert the relative water level into an absolute water level, a virtual staff gauge is introduced. The construction of the virtual staff gauge depends on whether a physical staff gauge exists at the monitoring station, resulting in two different scenarios.

For stations equipped with a physical staff gauge, several studies \cite{9359022, app13095614} utilize feature points on the gauge for perspective correction. This paper adopts a similar strategy. Specifically, a perspective transformation is performed using the four corner points of the staff gauge to compensate for the inconsistency between image pixel scale and real-world physical scale caused by perspective distortion. After transformation, each pixel corresponds to a uniform physical distance in the real world, thereby enabling more accurate water level estimation.

For stations without a physical staff gauge, surveying instruments such as total stations or RTK systems are required to measure a series of reference points along the direction of water level variation and obtain their real-world elevations. Based on the correspondence between image coordinates and surveyed reference points, the physical elevations of other pixels can then be estimated through interpolation. However, due to perspective effects and the non-uniform slope between calibration points, directly interpolated elevations inevitably contain estimation errors. To further improve water level estimation accuracy, some studies \cite{hess-27-4135-2023} employ 3D scanning techniques to reconstruct the entire riverbank terrain geometry.

Before performing the aforementioned processing steps, image registration is required. Owing to environmental disturbances such as wind-induced camera motion, the image captured at the current time is often misaligned with the reference image used during virtual staff gauge calibration. Therefore, the original camera pose must first be recovered. This can be achieved through feature matching algorithms, such as \cite{lindenberger2023lightglue}, or through camera pose estimation methods\cite{9740209} based on fiducial markers such as AprilTag or ArUco.

Finally, after obtaining the estimated water level, a water level prediction model\cite{hess-30-797-2026,ayus2023prediction} can be further incorporated to identify and filter potentially erroneous recognition results based on historical water level observations.

\textbf{Vision-based flow estimation:}
Vision-based river surface velocity estimation includes methods such as STIV (Space-Time Image Velocimetry) \cite{fujita2007development}, LSPIV (Large-Scale Particle Image Velocimetry) \cite{fujita1998large}, OTV (Optical Tracking Velocimetry) \cite{tauro2018optical}, and pixel-level feature tracking methods \cite{lemoing2024dense, karaev24cotracker3}. Fundamentally, all of these approaches estimate flow velocity by tracking the motion of surface features on the river.

The final measurement accuracy is influenced by several factors. The first factor is the accuracy of river surface feature tracking itself. The performance of vision-based tracking methods strongly depends on the visibility and stability of surface features, river flow conditions, as well as environmental factors such as illumination changes and weather conditions.

In this work, the STIV method is adopted for surface velocity estimation. However, relying on a single measurement line is often insufficient to achieve reliable and stable results. Therefore, this paper approaches the problem from a statistical perspective by modeling the lateral correlation of river surface velocities across the river cross-section. Specifically, the cross-section is divided into multiple segments, and multiple measurement lines are defined within each segment. The velocity corresponding to the maximum probability density is then selected as the representative velocity of that segment, thereby improving robustness against noisy or unreliable measurements.

The second factor affecting accuracy is the transformation from image coordinates to real-world physical coordinates. Existing solutions can generally be categorized into two groups. The first estimates flow velocity using pre-calibrated camera intrinsic and extrinsic parameters, while the second performs on-site calibration using ground control points. The former is easier to deploy in practical applications, whereas the latter typically provides higher measurement accuracy.

The third factor is that vision-based approaches can only measure river surface velocity rather than directly estimate river discharge. To calculate discharge, the measured surface velocity must be converted into depth-averaged velocity using a correction coefficient. Some studies \cite{IchiroFujita2024, OMORI20241122,yoshimura2024numerical} estimate this coefficient using methods such as the maximum entropy method. However, obtaining accurate discharge estimates still requires calibration under low-, medium-, and high-water-level conditions using ground-truth discharge measurements. In practice, this calibration process is often labor-intensive, operationally complex, and resource-consuming.

Finally, a stage–discharge relationship curve can be established based on the measured water level and discharge data. By fitting the water level–discharge relationship, abnormal discharge measurements can be identified and filtered out. Furthermore, the relationship curve can be calibrated and refined using ground-truth water level–discharge observation pairs, thereby improving the accuracy and reliability of discharge estimation.

\section{Method}
\subsection{Vision-based water level recognition}
\begin{figure}[htbp]
    \centering
    \includegraphics[width=\linewidth]{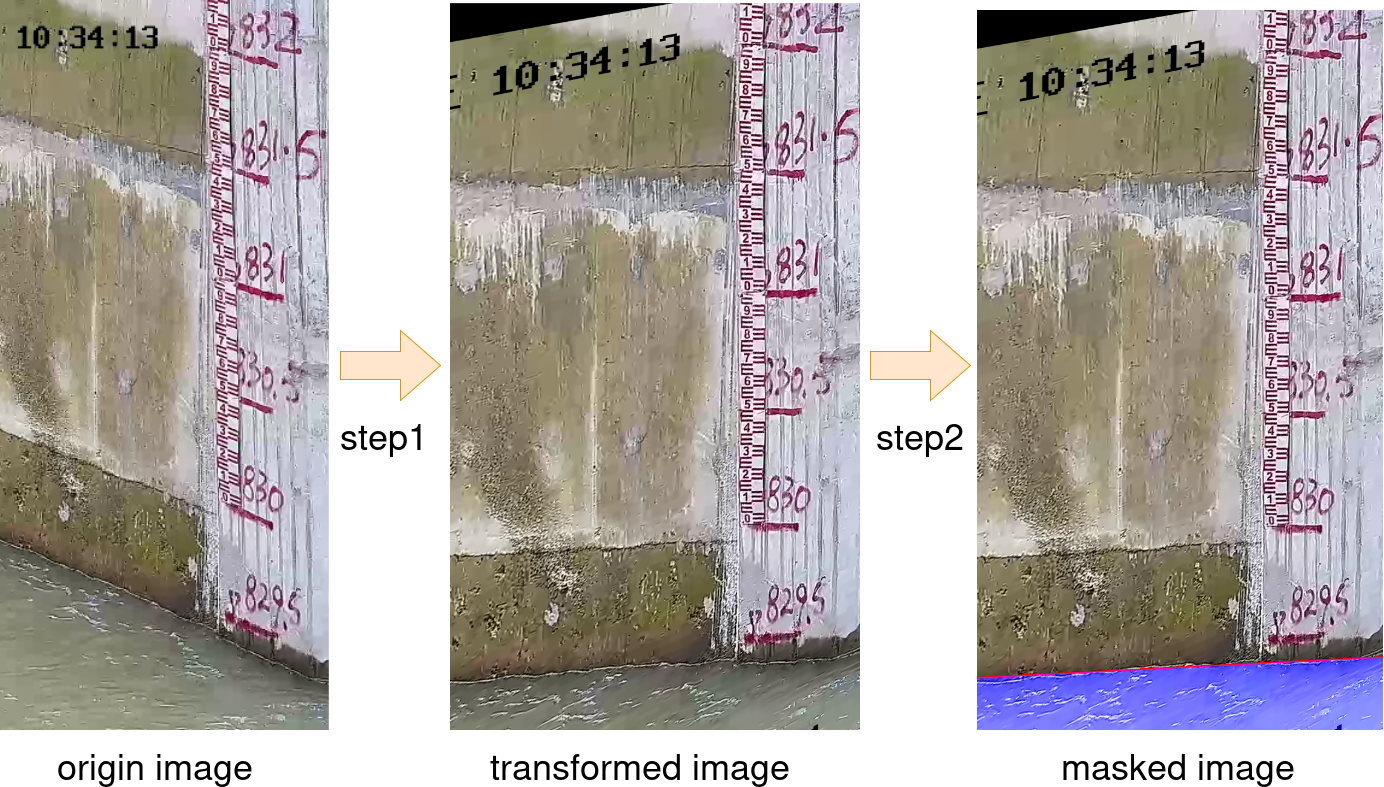}
    \caption{Overview of the proposed vision-based water level recognition pipeline. Step 1 performs image coordinate transformation and perspective rectification to align the current image with the reference virtual staff gauge image under a unified physical scale. Step 2 applies prior-guided semantic segmentation and boundary-slope validation to extract reliable water body regions and filter low-quality segmentation results.}
    \label{water_level_pipeline}
\end{figure}

The overall water level recognition pipeline is illustrated in Fig.~\ref{water_level_pipeline}. After the camera is dispatched to the designated monitoring location and the raw image is acquired, Step 1 performs image coordinate transformation, where the current image is mapped onto a reference virtual staff gauge image. This process includes camera pose correction using LightGlue\cite{lindenberger2023lightglue} and perspective correction based on four reference points on the physical staff gauge, ensuring that each pixel in the transformed image corresponds to a consistent real-world physical scale.

\begin{figure}[htbp]
    \centering
    \includegraphics[width=\linewidth]{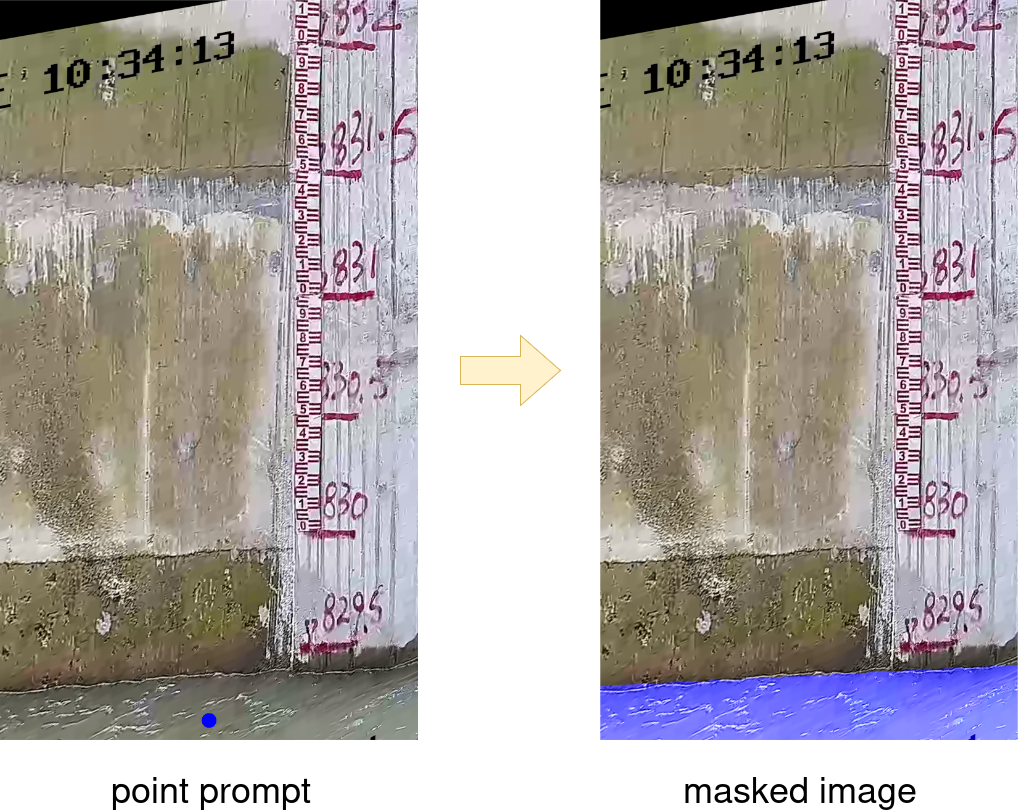}
    \caption{Overview of the proposed vision-based water level recognition pipeline. Step 1 performs image coordinate transformation and perspective rectification to align the current image with the reference virtual staff gauge image under a unified physical scale. Step 2 applies prior-guided semantic segmentation and boundary-slope validation to extract reliable water body regions and filter low-quality segmentation results.}
    \label{segment}
\end{figure}

Step 2 applies a prior-knowledge-guided semantic segmentation algorithm for water body extraction, where the blue region represents the segmented water body. The red line denotes the physical boundary between the water surface and the riverbank slope. This physical boundary can be manually calibrated in advance, allowing its slope to be determined beforehand. The slope fitted from the segmented water boundary curve is then compared with the known boundary slope to evaluate the segmentation quality, thereby filtering out segmentation results with poor reliability.
We now describe in detail the prior-knowledge-guided semantic segmentation method and the corresponding filtering strategy, as illustrated in Fig.~\ref{segment}. In water level monitoring images, the water region is relatively constrained in spatial position. In this setting, regardless of how the water level changes, the water body consistently appears in the lower part of the image. Therefore, we can define a critical region threshold such that the water level will never fall below it.

Based on this assumption, a point can be automatically sampled from the image region below the predefined threshold. This point, together with the image, is then fed into an interactive segmentation model such as RITM\cite{reviving2021} or SAM2\cite{ravi2024sam2}, where point prompts are used to guide the model to extract the water body using its inherent visual characteristics.
For sites affected by illumination changes, reflections, or other disturbances, an additional iterative refinement strategy is introduced using a prior on the slope of the water–bank boundary (see the red solid line in Fig.~\ref{water_level_pipeline}). Starting from an initial segmentation, the water mask is iteratively refined until convergence. The final segmentation is then validated by comparing the estimated water–bank boundary with the physical prior slope. Results with large deviations are filtered out.

After obtaining the segmented mask, we describe how to automatically assign a quality score to each segmentation result and filter out low-quality outputs.

Let the predicted binary mask be:
\[
M \in \{0,1\}^{H \times W}.
\]

We first compute the column-wise water pixel density:
\[
y(x) = \sum_{i=1}^{H} M(i,x), \quad x = 1,2,\dots,W.
\]

Let:
\[
\mathbf{y} = [y(1), y(2), \dots, y(W)].
\]

A reference vertical alignment position is defined as:
\[
x_{\mathrm{ref}} = \frac{W}{2}.
\]

To compensate for the physical prior slope of the water–bank boundary, we apply a linear correction:
\[
\tilde{y}(x) = y(x) - s \cdot (x - x_{\mathrm{ref}}),
\]
where \( s \) denotes the known physical prior slope of the water–bank boundary.

The segmentation quality is then evaluated using the variance:
\[
\mathcal{Q} = \mathrm{Var}(\tilde{\mathbf{y}}).
\]

A lower value of \( \mathcal{Q} \) indicates higher consistency with the physical prior, and thus a higher-quality segmentation result. Therefore, segmentation outputs with large post-correction variance are filtered out.

\subsection{Vision-based flow estimation}
\begin{figure}[htbp]
    \centering
    \includegraphics[width=\linewidth]{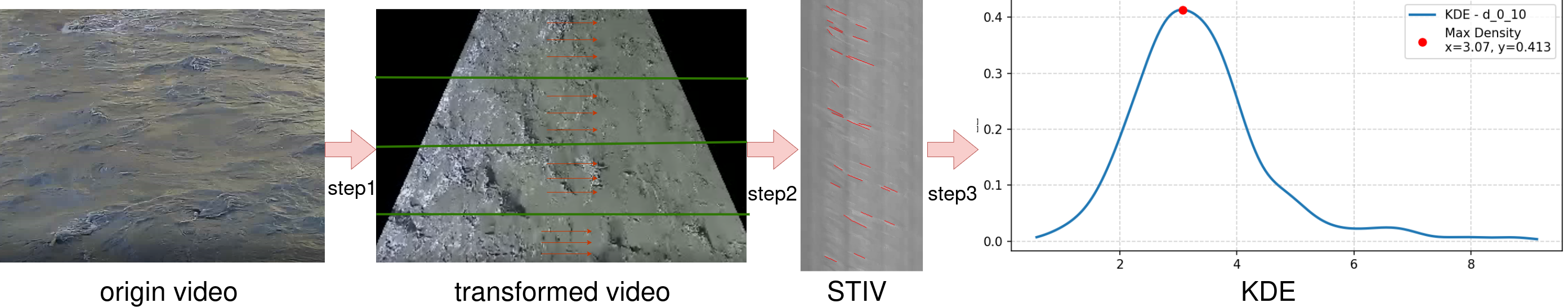}
    \caption{Workflow of vision-based river surface velocity estimation: top-down video transformation, cross-sectional segmentation, STI generation, and velocity estimation using Gaussian kernel density.}
    \label{flow_pipeline}
\end{figure}

The overall workflow for vision-based river surface velocity estimation is illustrated in Figure~\ref{flow_pipeline}. First, using the camera's intrinsic and extrinsic parameters, the original river surface video is transformed into a top-down view. This transformation mitigates the inconsistency of physical scales caused by perspective, ensuring that each pixel in the transformed video represents a fixed physical distance, denoted as $s$ meters per pixel. Next, the transformed video is segmented along the cross-section, and the Space-Time Image Velocimetry (STIV) method is applied to estimate the flow velocity for each segment. To enhance the robustness and accuracy of the results, a space-time image (STI) is generated for each pixel along the cross-sectional direction, capturing the flow velocity at the corresponding location. For each segment, Gaussian kernel density estimation (KDE) is used to determine the most probable maximum velocity, yielding the final flow velocity for that segment. The detailed processing procedure is described below.
In the top-down view transformation, we leverage the intrinsic and extrinsic parameters of the camera. Modern industrial PTZ cameras provide interfaces to obtain $f_x$, $f_y$, and PTZ parameters. However, due to factors such as camera installation, it is necessary to calibrate the pitch angle on-site. Specifically, along the cross-section of the river to be monitored, we measure the coordinates of a distant reference point and the camera to calculate the true pitch angle, then compare it with the pitch angle obtained from the camera to perform a correction. An additional reference point is measured for verification. Finally, the original video is transformed into a top-down perspective video using the current water level, intrinsic parameters, and the corrected extrinsic parameters.

After obtaining the top-down view video, a space-time image (STI) is generated for each pixel along the cross-section. Accurately identifying the texture orientation in the STI is critical for precise flow velocity estimation. Some approaches use Fourier transform~\cite{fujita2019application,wang2024adaptive} or deep learning~\cite{watanabe2021improving} to detect the texture angle in the STI. In our practice, we found that directly detecting the texture lines in the STI using LSD~\cite{von2008lsd} or DeepLSD~\cite{pautrat2023deeplsd} and determining their orientations works effectively. Line segments with obviously incorrect angles are filtered out using an angular threshold. Each remaining line segment corresponds to a flow velocity. After obtaining the velocities for all pixels in a segment, Gaussian kernel density estimation is applied to determine the most probable maximum velocity representing that segment. This method also returns a confidence score, which is compared with a preset threshold to decide whether to retain the measured segment velocity.

\begin{figure}[htbp]
    \centering
    \includegraphics[width=\linewidth]{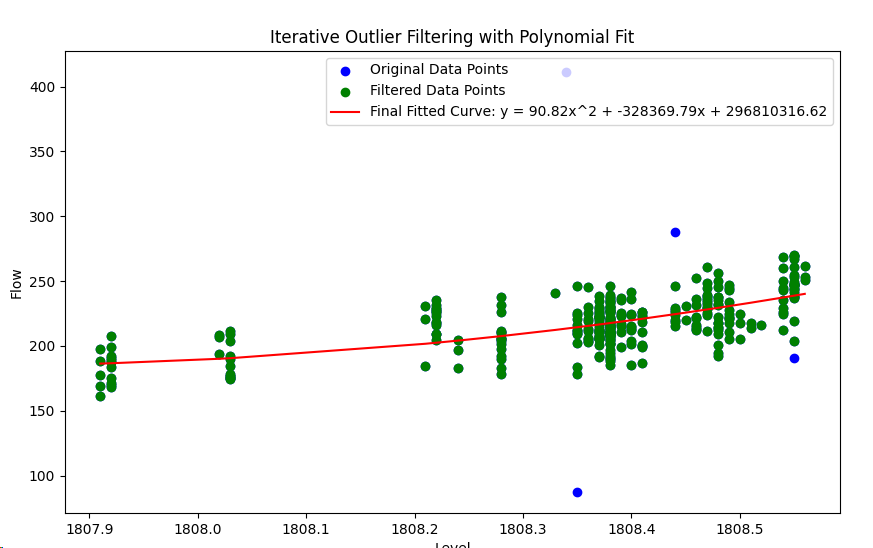}
    \caption{Illustration of the data cleaning and outlier removal procedure. The red curve represents the fitted quadratic polynomial, blue points indicate filtered outliers, and green points denote the inliers that satisfy the fitting criteria.}
    \label{water_level_flow}
\end{figure}

After obtaining the segmented water level and flow data, the next step is to perform data cleaning and outlier removal based on the water–flow relationship curve. Specifically, for each segment, an iterative fitting with a quadratic polynomial
\[
Q = aH^2 + bH + c
\]
is applied, and data points with large residuals are removed according to a relative error threshold (e.g., 20\%). If too many outliers are present, the fitted curve may exhibit negative derivatives within the water level range. In this case, a linear polynomial can be used for initial fitting to remove obvious outliers, followed by quadratic fitting. As shown in Fig.~\ref{water_level_flow}, the red curve represents the fitted quadratic polynomial, the blue points indicate the filtered outliers, and the green points denote the inliers that meet the criteria.
After removing outliers and filtering flow measurements with low KDE confidence, some flow segments may have missing values at certain time steps. To address this, we model the cross-sectional flow distribution using a Markov Random Field (MRF). Each flow segment (e.g., $d_{20}, d_{40}$) is treated as a node in the graph. For $N$ spatial segments, the algorithm first learns the joint flow distribution between any two segments $i$ and $j$ from historical data.

Let the historical flow matrix be $\mathbf{H} \in \mathbb{R}^{T \times N}$, where $T$ is the number of time steps. For a segment pair $(i,j)$, the sample set where both segments have nonzero flow is:
\[
\mathcal{D}_{ij} = \{ (\mathbf{h}_t^{(i)}, \mathbf{h}_t^{(j)}) \mid \mathbf{h}_t^{(i)} > 0 \land \mathbf{h}_t^{(j)} > 0, \forall t \in [1, T] \}.
\]

The joint probability density function (PDF) is estimated using Gaussian Kernel Density Estimation (KDE):
\[
P(v_i, v_j) \approx \text{KDE}_{ij}(v_i, v_j) = \frac{1}{|\mathcal{D}_{ij}| h^2} \sum_{(\tilde{v}_i, \tilde{v}_j) \in \mathcal{D}_{ij}} K\left(\frac{v_i - \tilde{v}_i}{h}, \frac{v_j - \tilde{v}_j}{h}\right),
\]
where $K(\cdot)$ is the Gaussian kernel, and the bandwidth $h$ is adaptively set based on the sample standard deviation (in code, $h = 0.2 \times \sigma$). These $\text{KDE}_{ij}$ functions form the pairwise potentials in the graph.

To impute a missing flow vector $\mathbf{v} = [v_1, v_2, \dots, v_N]$, a global energy function $E(\mathbf{v})$ is defined to measure the compatibility of the current flow configuration with historical joint distributions. For all segment pairs $\mathcal{M}$ with "one missing, one observed" values ($v_i=0, v_j>0$ or vice versa), the energy is defined as the sum of negative log-likelihoods:
\[
E(\mathbf{v}) = - \sum_{(i, j) \in \mathcal{M}} \log P(v_i, v_j) = - \sum_{(i, j) \in \mathcal{M}} \log \left( \text{KDE}_{ij}(v_i, v_j) \right).
\]
Physically, a lower energy indicates that the flow configuration is more probable according to historical statistics, reflecting the spatial distribution of flow along the cross-section.

Imputing missing values is formulated as a constrained nonlinear optimization problem:
\[
\mathbf{v}^* = \arg \min_{\mathbf{v}} E(\mathbf{v})
\]
with constraints:

\begin{itemize}
    \item \textbf{Observed values fixed:} For originally nonzero observations $v_k^{\text{obs}}$, keep their values:
    \[
    v_k = v_k^{\text{obs}}, \quad \forall k \text{ where } v_k^{\text{obs}} > 0
    \]
    \item \textbf{Physical non-negativity:} All flow values must be positive:
    \[
    v_m \geq \epsilon, \quad \forall m \text{ where } v_m^{\text{obs}} = 0, \quad (\epsilon = 10^{-6})
    \]
\end{itemize}

\textbf{Initialization:} Missing values are initialized to the mean of known nonzero flows to accelerate convergence:
\[
v_m^{(0)} = \text{mean}(\{ v_k^{\text{obs}} \mid v_k^{\text{obs}} > 0 \}), \quad \forall m \text{ where } v_m^{\text{obs}} = 0
\]

\begin{figure}[htbp]
    \centering
    \includegraphics[width=\linewidth]{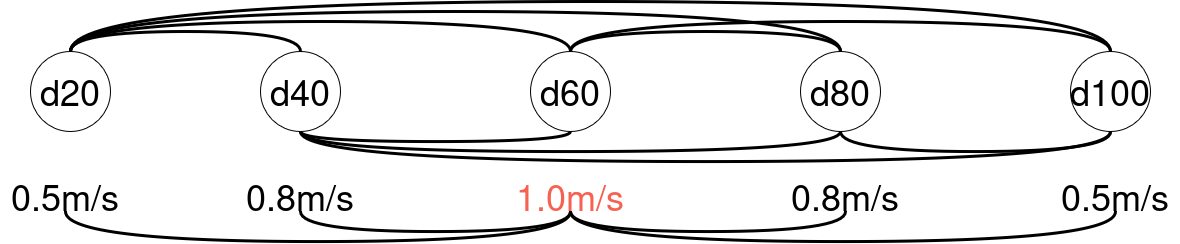}
    \caption{Illustration of the Markov Random Field (MRF) based flow imputation. Each segment (d20, d40, d60, d80, d100) is a node; the missing flow at d60 is inferred from the known flows of other segments using pairwise KDE potentials. Red font indicates the imputed value.}
    \label{mrf}
\end{figure}

The L-BFGS-B algorithm is then used to iteratively minimize $E(\mathbf{v})$, producing the physically consistent imputed flow vector $\mathbf{v}^*$.  
As illustrated in Fig.~\ref{mrf}, segments $d_{20}, d_{40}, d_{60}, d_{80}, and d_{100}$ represent different spatial positions. Here, the flow at $d_{60}$ is missing, while $d_{20}$ and $d_{100}$ are 0.5 m/s, and $d_{40}$ and $d_{80}$ are 0.8 m/s. Using the learned pairwise relationships, the missing $d_{60}$ flow is inferred as 1.0 m/s (red font indicates the imputed value).
After imputing the missing segment flow values, the flow for each segment is computed as the product of the segment velocity and segment area, and the total flow is obtained by summing all segments. A quadratic polynomial is then fitted to the total flow versus water level, and outliers are removed. Finally, the preliminary fitted curve $\hat{Q}(H)$ is calibrated using three known true flow measurements at low, medium, and high water levels $(H_j^*, Q_j^*), j=1,2,3$.

A Radial Basis Function (RBF) correction model is constructed in the logarithmic space by defining a correction factor $\alpha(H)$ such that
\[
Q_{\text{cal}}(H) = \hat{Q}(H) \cdot \alpha(H).
\]
In log-space, this is expressed as
\[
\log Q_{\text{cal}}(H) = \log \hat{Q}(H) + \sum_{j=1}^{3} w_j \phi_j(H),
\]
where the Gaussian kernel functions $\phi_j(H)$ are defined as
\[
\phi_j(H) = \exp\left( -\frac{(H - H_j^*)^2}{2\sigma_j^2} \right),
\]
with bandwidth $\sigma_j$ adaptively determined based on the local spacing of the known control points. The weight vector $\mathbf{w} = [w_1, w_2, w_3]^T$ is obtained by solving the regularized linear system
\[
(\mathbf{\Phi} + \lambda \mathbf{I}) \mathbf{w} = \mathbf{y},
\]
where $\mathbf{\Phi}_{ij} = \phi_j(H_i^*)$, and the target vector is $\mathbf{y}_i = \log Q_i^* - \log \hat{Q}(H_i^*)$.

The final calibrated water level–flow relationship is thus
\[
Q_{\text{final}}(H) = \hat{Q}(H) \cdot \exp\left( \sum_{j=1}^{3} w_j \exp\left( -\frac{(H - H_j^*)^2}{2\sigma_j^2} \right) \right).
\]

This approach ensures that the computed flow maintains the physical trend (monotonicity of the power-law) while accurately passing through the known control flow points.
\section{Experiments}
We applied a prior-knowledge-based semantic segmentation algorithm to various water-level monitoring sites. This algorithm can effectively segment the water body and calculate water level without the need for additional training data. Typically, when a staff gauge is present, the water-level recognition accuracy is around 5 cm, whereas in the absence of a staff gauge, the error increases to approximately 10 cm. To achieve high accuracy, it is crucial to precompute the physical distance represented by each pixel.

In addition, our vision-based flow estimation method exhibits an error of about 10\%. The accuracy of vision-based flow estimation, however, is influenced by several factors, including site selection, camera installation height, and angle correction precision. Among these, site selection is particularly critical: choosing a location with rich surface features and smooth flow significantly improves both accuracy and stability. Camera height also plays an important role, as the farthest point's top-down viewing angle should not be too small. Moreover, camera height is constrained by the texture angle: in STI images, when the texture angle is too small, the measured flow velocity is highly sensitive to changes in texture angle, reducing accuracy.

\section{Conclusion}

We presented an integrated framework for vision-based water level and flow estimation, combining state-of-the-art computer vision models with robust statistical modeling. By leveraging physical priors, such as the geometric stability of shorelines and the spatial correlation of cross-sectional flow, the system minimizes the need for site-specific training data. 

Our prior-knowledge-guided segmentation approach, using models like SAM2 and RITM, allows immediate deployment across diverse monitoring stations, achieving water level accuracy within 5--10 cm. For flow estimation, integrating STIV with Gaussian Kernel Density Estimation and Markov Random Field-based imputation provides robustness against noisy measurements and missing data. Additionally, RBF-based calibration ensures that the derived stage–discharge relationships remain accurate and physically consistent.

\bibliography{biblio.bib}

\end{document}